\documentclass{article}
\usepackage{PRIMEarxiv}
\usepackage[utf8]{inputenc} % allow utf-8 input
\usepackage[T1]{fontenc}    % use 8-bit T1 fonts
\usepackage{hyperref}       % hyperlinks
\usepackage{url}            % simple URL typesetting
\usepackage{booktabs}       % professional-quality tables
\usepackage{amsfonts}       % blackboard math symbols
\usepackage{nicefrac}       % compact symbols for 1/2, etc.
\usepackage{microtype}      % microtypography
\usepackage{fancyhdr}       % header
\usepackage{graphicx}
\usepackage[frozencache, cachedir=.]{minted}
\graphicspath{{media/}}     % organize your images and other figures under media/ folder
\usepackage{xcolor}
\usepackage{tabularx}
\usepackage{times}
\usepackage{latexsym}
\usepackage{rotating}
\usepackage{amssymb}
\usepackage{array}  
\usepackage{longtable}
\usepackage{pgfplots}
\usepgfplotslibrary{groupplots}
\pgfplotsset{compat=1.18}

\setminted{
  breaklines=true,
  breaksymbolleft={},
  breaksymbolright={},
  frame=lines,
  framesep=2mm,
  baselinestretch=1.2,
  fontsize=\footnotesize,
  encoding=utf8
}

\title{Brasilidade: Measuring Brazilian Identity for Large Language Models}
\author{ José Roberto Homeli da Silva, Felipe Jorge do Nascimento, Lucas Coutinho Freitas,\\ \textbf{William Alberto Cruz Castañeda, Marcellus Amadeus} \\
  SoberanIA \\
  \texttt{\{homeli, felipe, lucas, william, marcellus\}@soberania.ai} \\
}

\begin{document}
\maketitle

\begin{abstract}
This work presents a new benchmark that aims to address the limitations of current methods for evaluating LLMs in Brazilian Portuguese. We introduced \textit{Brasilidade}, an assessment methodology comprised of standardized tasks distributed across four different domains, and we are making available the tasks from its first domain, which address nuances of the language, grammar, and variations of Brazilian Portuguese. The results show that model size, by itself, does not determine performance; several medium-sized models outperformed larger models in the protocols evaluated. Specific models for Portuguese, in general, did not perform above average, despite their targeted training. Different task levels revealed significant nuances: while some tasks proved easy for most models, others posed extreme challenges, and some yielded mixed results, with wide variations in performance. This clearly illustrates the need for benchmarks that address the limitations of the models in Brazilian Portuguese, allowing future training cycles to seek to overcome them. Project repository \href{https://github.com/soberania-oficial/brasilidade}{https://github.com/soberania-oficial/brasilidade}
\end{abstract}

% keywords can be removed
\keywords{Benchmark \and Open-source Large Language Models \and Brazilian Portuguese}

\section{Introduction}
The word \textit{Brasilidade} can be translated literally as \textit{Brazilianness}, the unique national identity of Brazil. With the current widespread adoption of Artificial Intelligence (AI) systems by Brazilian users \cite{cruz-castanedaLargeLanguagesModels2025}, Large Language Models (LLMs) need not only a working knowledge of Brazil's specificities, but rather grasp the complexities of \textit{Brazilianness} as a concept reflected across all fields where AI and the largest country in Latin America intersect. The lack of availability of cutting-edge technologies adapted to the necessities and realities of Brazil's population cannot be ignored in today's landscape. However, since capturing the entire essence of a country as diverse and geographically large as Brazil is virtually impossible, it is no surprise that even the most advanced models fail at seemingly simple tasks that require specific knowledge that has not yet been explored by their developers \cite{piresSabiaPortugueseLarge2023}. Attempts to fill the gap in LLMs performance between different languages have been made by the community \cite{ruderXTREMERMoreChallenging2021} and advancements in this field indicate that benchmarks are built to challenge current models and their biases as much as these models must be built to adapt to the multiple perspectives and world views reflected in the most challenging benchmarks of their time \cite{srivastavaImitationGameQuantifying2023,changSurveyEvaluationLarge2024,ottMappingGlobalDynamics2022}. As such, the \textit{Brasilidade} benchmark's most valuable asset is the challenge it poses to models with a vague comprehension of Brazil, shifting more focus to the development of new models that not only understand, but are part of Brazil's national identity. \textit{Brasilidade} is part of the SoberanIA\footnote{https://www.soberania.ai} research initiative, which aims to prioritize Brazilian sovereignty in AI as a first-class design requirement at all levels. Thus, the first version is a suite of evaluation tasks over four domains:

\begin{itemize}
  \item \textit{Nuances of language, grammar, and variations of Brazilian Portuguese.}
  \item \textit{Culture and general knowledge of Brazil.}
  \item\textit{Performance in tasks applied to the Brazilian context.}
  \item \textit{Safety and bias analysis in the Brazilian context.}
\end{itemize}

Here we present the complete benchmark methodology, but only the first domain will be made publicly available immediately and discussed in the results section. Our main contributions are as follows:

\begin{itemize}
  \item A standardized evaluation methodology for LLMs capable of responding in Brazilian Portuguese that ensures reproducibility of results as per academic standards \cite{stoddenImplementingReproducibleResearch2018};
  \item 27 publicly available tasks of the domain \textit{nuances of language, grammar, and variations of Brazilian Portuguese}, of which six are specifically focused on language variation and 21 are focused on Portuguese-language grammar;
  \item Eight unique datasets, curated and reviewed by Brazilian linguists, served as the basis for the 27 tasks in the first domain. 
\end{itemize}

\section{Related Work}
There have been a great number of recent efforts from the community to solve the problems detailed above. To the best of our knowledge, however, there are no standardized benchmarks with datasets that were not translated and were built and curated by native speaker linguists with the purpose of evaluating models on the nuances of Brazilian language, culture, education, institutions, and identities in the same suite.
\subsection{Portuguese-Language Benchmarks}
Existing benchmarks in Portuguese are limited, lacking cultural specificity for Brazil, linguistic and grammatical challenges, or a careful production pipeline that validates dataset entries using native speakers. %\subsubsection{ALBA}
\textbf{ALBA} \cite{vieiraALBAEuropeanPortuguese2026} is a benchmark designed to assess LLMs proficiency in linguistic tasks for European Portuguese across eight dimensions. It was manually constructed by language experts, and its evaluation is performed by an LLM-as-a-judge framework. %\subsubsection{Tuguesice-PT}
 \textbf{Tuguesice-PT} \cite{silvaCLARINPTLDBOpenLLM2026} is a benchmark that tests specific cultural knowledge of Portugal. The questions were devised to implicitly assume full familiarity with Portuguese culture, as if in an exchange between two people born and raised in Portugal, and cover a variety of topics, such as society, geography, history, politics, gastronomy, and sports. %\subsubsection{PoETa suite}
The \textbf{PoETa} suite is the standard benchmark for LLMs in Brazilian Portuguese \cite{almeidaPoETaV2More2025}. It has two versions, with 9 and 44 tasks respectively. Most of the tasks were translated from English.

\subsection{Other Native Benchmarks}
As interest in multilingual and culturally grounded evaluation grows, several benchmarks have been created for underrepresented languages. These efforts adapt large-scale evaluation principles to specific linguistic and regional contexts, ensuring relevance, fairness, and coverage beyond English-dominant tasks. %\subsubsection{MERA} 
\textbf{MERA} \cite{fenogenovaMERAComprehensiveLLM2024} is a comprehensive benchmark for evaluating LLMs in Russian, modeled after large-scale English-language evaluations. It covers 21 evaluation tasks for generative models in 11 skill domains and is designed as a black-box test to ensure the exclusion of data leakage. It presents a methodology to evaluate foundation models in zero and few-shot fixed instruction that can be extended to other modalities. MERA's methodology is relevant for evaluations in underrepresented languages that capture linguistic nuance and regional context. %\subsubsection{IrokoBench}
\textbf{IrokoBench} \cite{adelaniIrokoBenchNewBenchmark2024} introduces a multilingual benchmark covering 17 low-resource African languages. It includes three types of task: natural language inference, mathematical reasoning, and multi-choice knowledge-based Q\&A. All tasks are human-translated and curated to ensure cultural and linguistic validity. This is a prime example of how multilingual evaluation can be extended to cover typologically diverse, underrepresented languages. %\subsubsection{ITALIC}
\textbf{ITALIC} \cite{sevesoITALICItalianCultureAware2025} is a large-scale benchmark dataset of 10,000 multiple-choice questions designed to evaluate the natural language understanding of the Italian language and culture. It spans 12 domains, exploiting public tests to score domain experts in real-world scenarios.

\section{Data}
The construction of the dataset was based on a hybrid methodology that prioritized both linguistic quality and technical scalability. The workflow was conducted by experts in computational linguistics, ensuring both methodological rigor and appropriate cultural representation throughout at all stages of validation. Some datasets have multiple fields, and different combinations of those fields were used to build tasks, adding up to a total of more than 4,000 unique entries and almost 15,000 total entries throughout the benchmark. In addition, it organizes unstructured data in rigorous formats, such as terminology tables, grammar annotation, and Q\&A pairs. For data generation, we adopted two complementary approaches:

\begin{itemize}
    \item \textbf{Human generation and curation}: applied to tasks involving high cultural sensitivity, complex logical reasoning, and specific grammatical nuances. In these instances, the intuition of the native-speaker and manual authoring were deemed irreplaceable for handling semantic challenges.
    \item \textbf{Assisted synthetic generation}: used for data augmentation and creation of rule-based examples and raw source text for data transformation. This process incorporated few-shot learning techniques, providing the models with manually curated examples that served as reference templates for large-scale generation.
\end{itemize}

\subsection{Quality assurance}
Beyond the technical expertise of linguists and the application of modern prompt-engineering strategies, all synthetically generated material underwent a rigorous human review phase. This audit was conducted through representative sampling or exhaustive inspection, ensuring the correction of potential hallucinations and deviations from cultural appropriateness and standard linguistic norms.

\subsection{Processing and curation pipeline}
The formulation of datasets based on general domain knowledge followed certain guidelines:

\begin{itemize}
    \item Curated primary sources covering a wide range of materials from textbooks, web pages, and newspaper articles to scientific papers.
    \item Raw data extraction using web scraping and optical character recognition techniques.
    \item Normalized information in plain text files, structured line-by-line.
    \item Factual Q\&A pairs created via assisted generation.
    \item The resulting raw data were re-segmented into specific domains of relevance to ensure a balanced distribution for benchmarking.
    \item For structured tasks, such as multiple-choice questions, conflicting alternatives (distractors) were either generated or sampled from the ground-truth answers of other instances to increase evaluation complexity.
    \item Entries that failed to meet the strict inclusion criteria or lacked evaluation coherence were not discarded; instead, they were repurposed and redirected to enrich post-training datasets for future models.
\end{itemize}

More specific information on dataset construction, curation and structure can be found on Appendix \ref{sec:appendix_A}.

\section{Benchmark}
\label{sec:benchmark}
\textit{Brasilidade} spans over four domains and is structured in multiple layers, following a pipeline that starts from two distinct inference modes depending on output type (generation and probability-based), and is further divided into classification and open-ended questions, which are in turn evaluated according to six different task templates.

\subsection{Domains}
\textit{Brasilidade} aims to assess model performances in tasks that encompass multiple dimensions and use cases for the Brazilian diaspora. For that reason, we have established four domains that reflect the most important aspects of national identity and particularities to set the benchmark's direct evaluation priorities. Brazilian identity in this context is then linked to representation and identification \cite{carneirorodriguesIdentidadesBrasileirasComposicoes2014}, the pillars upon which the four domains of the benchmark were constructed.

\begin{enumerate}
    \item \textbf{Nuances of language, grammar, and variations of Brazilian Portuguese} covers both the normative grammar of the Portuguese language and the sociolinguistic and cultural aspects of spoken Portuguese in Brazil. The Grammar subgroup focuses on tasks that require morphological and syntactic knowledge: word classification, verb conjugation, morpheme identification, formation of augmentatives and diminutives, correction of grammatical errors, and categorization of the type of error committed. The Language Variations subgroup assesses knowledge of colloquial and regional expressions. All tasks and datasets of this domain were made public.
    \item \textbf{Culture and general knowledge of Brazil} assesses factual knowledge about Brazil in its various dimensions. It covers areas such as national history, cuisine, traditions, heritage, tourism, and specific regional content. Tasks, datasets and results related to this domain will be released soon.
    \item \textbf{Performance in tasks applied to the Brazilian context} covers general knowledge and reasoning within the educational and institutional context of Brazil. Tasks, datasets and results related to this domain will be released soon.
    \item \textbf{Safety and bias analysis in the Brazilian context} evaluates the model's ability to identify potentially harmful or biased content in a Brazilian context. Tasks, datasets and results related to this domain will be released soon.
\end{enumerate}

\subsection{Evaluation Framework}
\begin{figure*}[h]
    \centering    \includegraphics[width=\textwidth,trim={0 350 0 350}, clip]{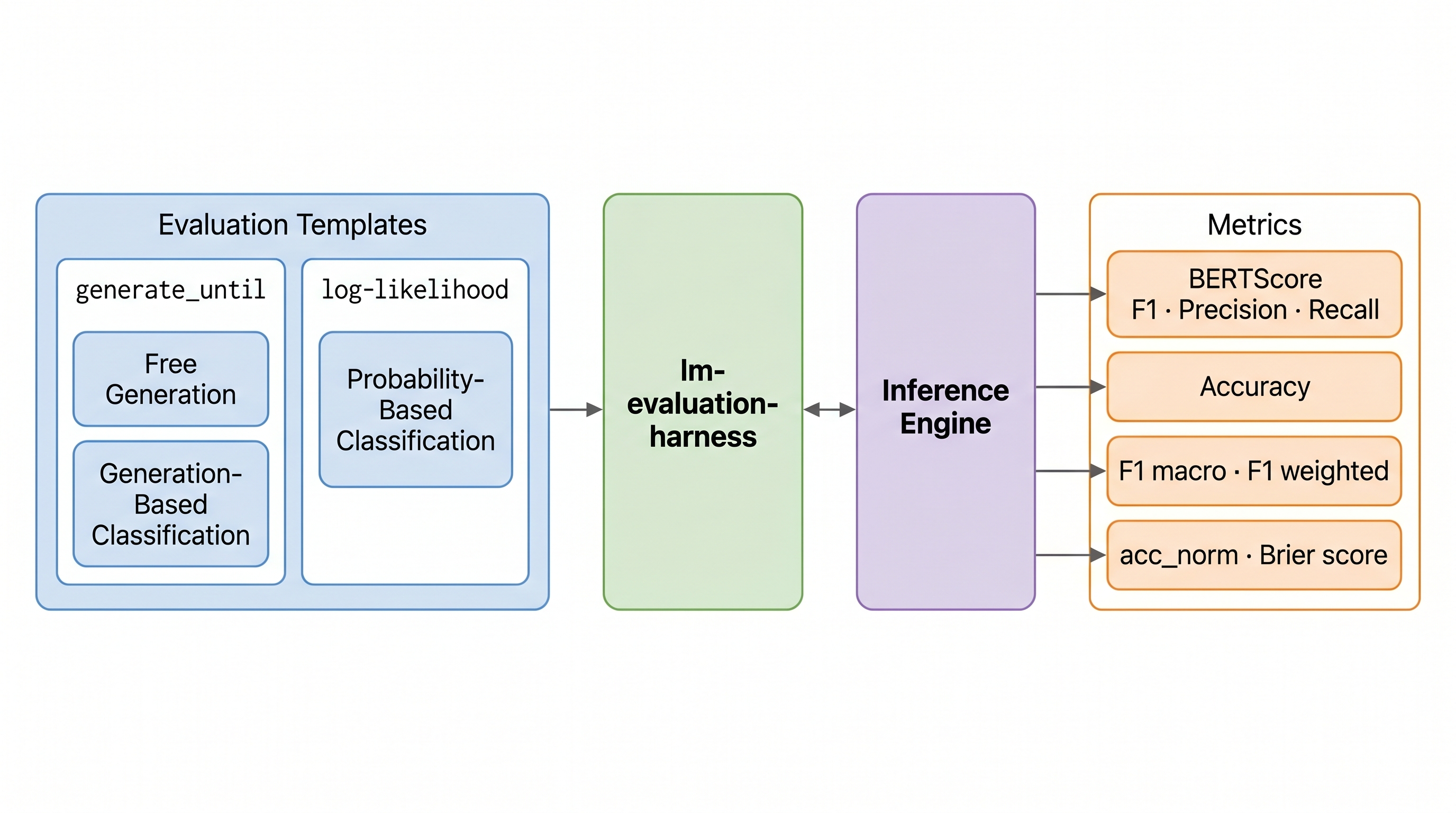}
    \caption{Overview of the evaluation pipeline, illustrating the standardized workflow from task configuration evaluation base templates to orchestration and optimized inference engine, culminating in downstream statistical and semantic evaluation metrics.}
\label{fig:evaluation_pipeline}
\end{figure*}

All tasks in \textit{Brasilidade} were built using the LM Evaluation Harness framework by Eleuther AI \cite{gaoFrameworkFewshotLanguage2021}, one of the most widely adopted frameworks for evaluating LLMs in both academic and industry settings. The entire evaluation pipeline is depicted in Figure  \ref{fig:evaluation_pipeline}, structured as a streamlined, high-throughput workflow designed to ensure rigorous reproducibility and guarantee that all tasks are applied identically and uniformly across all evaluated models. The base templates of each task type are extended by each specific task configuration and ingested by the LM Evaluation Harness framework, which manages concurrent execution and orchestrates requests using an inference engine that serves both model and tokenizer to ensure maximum token throughput and efficient memory utilization. Tasks built on this framework use one of these two types of output for evaluation:

\begin{itemize}
  \item \textbf{generate\_until} to generate a response from a prompt until it reaches a stop sequence or token limit.
  \item \textbf{loglikelihood} computes the log-probability that the model assigns to a continuation.
\end{itemize}

\subsection{Standardized Prompting}
The quality of the prompt provided to the model can drastically affect its responses and, consequently, its performance under evaluation \cite{weberMindInstructionsHolistic2023}. We standardized our prompts to maximize instruction-following. Depending on whether the task is based on generation or probability, the prompting style changes in order to better adapt to its respective output type.

\subsubsection{Generation Tasks}
All tasks follow the same prompt pattern. An instruction field positions the model as an expert in the relevant area, requests that the response always be in Brazilian Portuguese, and specifies the expected format (including, for classification tasks, an explicit listing of the possible options). Instructions are followed by examples (5 or 10, depending on the task type), which show the model the exact expected response format. The body of the prompt is structured around the markers \verb|'Question:'|  and \verb|'Answer:'|, with the latter acting as an induction anchor: the model completes from that point, with no room for preambles or explanations. Generation is stopped upon encountering two blank lines, which prevents the model from continuing beyond the relevant content. The maximum number of generated tokens varies by subtype: open Q\&A responses have more margin (100 tokens) than classification tasks (64 tokens) or direct generation (50 tokens).
\paragraph{Instructions.}
All generation tasks (free generation and generation-based classification) include standardized instructions in their prompts which describe the task, the model's expected behavior and explicitly state all other important information. The example below contains the instructions (translated into English) for the task \textit{pt\_conjugate}:

\begin{listing}[!h]
\begin{minted}{text}
You are a virtual assistant specialized in Portuguese grammar and must always answer in Brazilian Portuguese. Below are some examples of direct questions about verb conjugation and their respective answers. Your task is to generate the correct conjugation of a verb according to the question, including all necessary pronouns. Answer only the last question by conjugating the requested verb in the requested tense, following the same format as the examples. Answer only with the final content. Do not add explanations, justifications, introductions, or variations.
\end{minted}
    \caption{Instructions for the pt\_conjugate task, translated into English.}
    \label{lst:pt-conjugate-instructions}
\end{listing}

\paragraph{Few-shot examples.} All of the generation tasks receive few-shot examples with the rest of the prompt to ensure the correctness of the response format \cite{brownLanguageModelsAre2020}.

\subsubsection{Probability-Based Tasks}
These tasks do not use the description field and do not present examples to the model (zero-shot). The prompt is an incomplete sentence phrased naturally, without the question-and-answer structure. Each answer option is evaluated as an independent completion of that sentence, and the one with the highest log-probability is selected as the model's prediction.

\subsection{Task Types and Subtypes}
\label{sec:task_types_subtypes}
All benchmark tasks fall into one of three types, defined by how the model is evaluated. This choice determines the prompt format and the number of examples provided. All generation-based tasks that are not classification are Type 1, while classification tasks have 2 versions using the same datasets and values - one in the Q\&A format that is generation-based (Type 2 tasks) and one in the completions format that is probability-based (Type 3 tasks).
\subsubsection{\textbf{Type 1: Free Generation}}
The model receives an instruction prompt along with 10 few-shot examples and an open-ended question to produce a free-text response using \textit{generate\_until} as the output type. No options are presented: the answer key may be a sentence, a date, a proper noun, or any factual content. Evaluation is either done using semantic similarity metrics, which recognize paraphrases as correct, or by exact match in cases where the answer key is deterministic. Below, Listing \ref{lst:type1-prompt}, is an example of how type 1 tasks are shown to the model: instructions are given along with 10 few-shot examples, then the induction anchor with the question is shown so that the model answers in the expected format.

\begin{listing}[!h]
\begin{minted}{text}
[PROMPT INSTRUCTIONS (e.g. "You are a virtual assistant specialized in ...")]

[FEW-SHOT BLOCK WITH 10 EXAMPLES OF QUESTION AND ANSWER PAIRS]

Question: [QUESTION FROM THE DATASET]
Answer:
\end{minted}
    \caption{Prompt structure for Type 1 free generation, with ten few-shot question-and-answer examples.}
    \label{lst:type1-prompt}
\end{listing}

\begin{listing}[!h]
\begin{minted}{text}
[PROMPT INSTRUCTIONS (e.g. "You are a virtual assistant specialized in ...")]

[FEW-SHOT BLOCK WITH 5 EXAMPLES OF QUESTIONS (EACH WITH EXPLICITLY WRITTEN OPTIONS) AND THEIR RESPECTIVE ANSWERS]

Question: [QUESTION FROM THE DATASET WITH EXPLICITLY WRITTEN OPTIONS]
Answer:
\end{minted}   
    \caption{Prompt example with an open-ended question to produce a free-text response.}
    \label{fig:json_code10}
\end{listing}

\subsubsection{\textbf{Type 2: Generation-Based Classification}}
The model receives an instruction prompt along with 5 few-shot examples and a direct question which informs it of the available options to generate the correct alternative exactly as it appears in the examples. This task type applies both to tasks with fixed classes (e.g. classification questions whose options never change) and to tasks with variable classes (e.g. multiple-choice questions whose alternatives change with each example). Below is an example of how Type 2 tasks are shown to the model: instructions are given along with 5 few-shot examples that include their respective options, then the induction anchor with the question and its options is shown so that the model answers in the expected format.

\begin{listing}[!h]
\begin{minted}{text}
[PROMPT INSTRUCTIONS (e.g. "You are a virtual assistant specialized in ...")]

[FEW-SHOT BLOCK WITH 5 EXAMPLES OF QUESTIONS (EACH WITH EXPLICITLY WRITTEN OPTIONS) AND THEIR RESPECTIVE ANSWERS]

Question: [QUESTION FROM THE DATASET WITH EXPLICITLY WRITTEN OPTIONS]
Answer:
\end{minted}
    \caption{Prompt structure for Type 2 generation-based classification, with five few-shot examples and explicit answer options.}
    \label{lst:type2-prompt}
\end{listing}

\subsubsection{\textbf{Type 3: Probability-Based Classification}}
The model receives no instructions, no examples and produces no text. Instead, the model calculates the log-probability of each answer option as a completion of an incomplete prompt. The prompt is a natural sentence that ends just before the answer, and each alternative is evaluated independently as a plausible continuation. This format completely eliminates the influence of format choices on the model's response. Below is an example of how type 3 tasks are shown to the model: a simple incomplete sentence that prompts the model to answer the dataset's question indirectly. This approach was chosen because probability-based tasks do not produce generated tokens, but rather calculate the log-probability over a given list of options. Thus, this type of task performs classification by choosing the string with the highest probability of completing the provided sentence.

\begin{listing}[!h]
\begin{minted}{text}
[INCOMPLETE SENTENCE]
\end{minted}
    \caption{Prompt structure for Type 3 probability-based classification, using an incomplete sentence.}
    \label{lst:type3-prompt}
\end{listing}

\subsubsection{\textbf{Classification Tasks Have 2 Versions}}
As mentioned above, Type 2 and Type 3 tasks are different approaches to the same classification problem. We have standardized prompting for generation and probability metrics to measure both the model's knowledge of the evaluated topic and its adherence to the expected response format. By creating two tasks that evaluate on the same specific knowledge but have very different formats, we aimed to produce results that directly showed the effect of inference mode and instruction-tuning on performance. Below is an example for the tasks \textit{pt\_word\_class\_gen} and \textit{pt\_word\_class\_prob}, translated into English:

\textbf{pt\_word\_class\_gen}
\begin{listing}[!h]
\begin{minted}{text}
[PROMPT INSTRUCTIONS ("You are a virtual assistant specialized in Portuguese grammar and...")]

[FEW-SHOT BLOCK WITH 5 EXAMPLES OF QUESTIONS (EACH WITH EXPLICITLY WRITTEN OPTIONS) AND THEIR RESPECTIVE ANSWERS
e.g.
"Question: Among the possible options ('noun', 'verb', 'adjective', 'adverb' and 'conjunction'), what is the morphological classification of the word 'menosprezar'?
Resposta: verbo"]

[Question: [QUESTION FROM THE DATASET WITH EXPLICITLY WRITTEN OPTIONS
Answer: 
e.g.
"Question: Among the possible options ('noun', 'verb', 'adjective', 'adverb' and 'conjunction'), what is the morphological classification of the word 'ramagem'?
Answer: "]
\end{minted}
    \caption{Generation-based morphological classification prompt for pt\_word\_class\_gen.}
    \label{lst:pt-word-class-gen-prompt}
\end{listing}
For the generation-based version of \textit{pt\_word\_class}, the model is given instructions and 5 examples, then asked the morphological classification of the word in question. 

\textbf{pt\_word\_class\_prob}
\begin{listing}[!h]
\begin{minted}{text}
"The morphological classification of the word 'ramagem' is "
\end{minted}
    \caption{Probability-based morphological classification prompt for pt\_word\_class\_prob.}
    \label{lst:pt-word-class-prob-prompt}
\end{listing}
For the probability-based version, the model calculates the log-probability of each string ('noun', 'verb', 'adjective', 'adverb' and 'conjunction') as a completion to the sentence-prompt and the most probable option is chosen as the model's answer.

\subsection{Metrics}
\label{sec:metrics}
Choosing the right metrics is one of the biggest challenges in evaluation due to the gap between automatic and human evaluation, metric reliability issues, and the need for better protocols \cite{changSurveyEvaluationLarge2024}. In an attempt to minimize possible noise, we have standardized the evaluation procedure by sorting the tasks into six different categories depending on inference mode and metric used. They are as follows:

\begin{enumerate}
\item \textbf{Open-Ended Q\&A Generation.}
Type 1 tasks where the model receives open-ended questions whose answer may differ slightly from the gold-standard. \textit{BERTScore F1} \cite{zhang*BERTScoreEvaluatingText2019} with the model BERTimbau Large \cite{souzaBERTimbauPretrainedBERT2020} used as the main metric to calculate the semantic similarity between the generated answer and the reference answer based on its embeddings.
\item \textbf{Direct Generation.}
Type 1 tasks where the model is prompted with a question whose generated answer must be exactly the same as the gold standard. The metric \textit{exact match} is used because each answer is either correct or incorrect, including its formatting.
\item \textbf{Generation-Based Classification with a Fixed Set of Options.}
Type 2 tasks where the options presented in each question are exactly the same. The metric \textit{f1\_weighted} is used to account for any imbalances in the distribution of options across the dataset.
\item \textbf{Generation-Based Classification with a Variable Set of Options.}
Type 2 tasks where the options presented in each question are never the same. The metric \textit{accuracy} is used because each answer is either correct or incorrect.
\item \textbf{Probability-Based Classification with a Fixed Set of Options.}
Type 3 tasks where the options presented in each question are exactly the same. The metric \textit{f1\_weighted} is used here to account for any imbalances in the distribution of options across the dataset.
\item \textbf{Probability-Based Classification with a Variable Set of Options.}
Type 3 tasks where the options presented in each question are never the same. The metric \textit{accuracy} is used because each answer is either correct or incorrect.
\end{enumerate}
More specific information about each task can be found in Appendix \ref{sec:appendix_B}.

\section{Results}
The set of models selected for evaluation was strategically assembled to cover a wide range of parameter landscapes and diversity of providers. The list includes 19 models ranging from 2B to 70B parameters, which allows us to investigate the correlation between model scaling and its proficiency across \textit{log-likelihood} and generation tasks. Furthermore, the inclusion of models from leading vendors such as Google \cite{teamGemmaOpenModels2024}, NVIDIA \cite{nvidiaNVIDIANemotron32025}, MistralAI \cite{jiangMistral7B2023}, Qwen \cite{teamQwen35NativeMultimodal2026}, and IBM \cite{IBMGraniteTechnical2024}, alongside Brazilian Portuguese-specific models like Tucano \cite{correaTucano2Cool2026} and Sabia \cite{piresSabiaPortugueseLarge2023}, aims to mitigate any specific architectural or training bias.

The results clearly show a pronounced gap between performance on few-shot generation (Gen) tasks and 0-shot log-likelihood (Prob) tasks, as seen in Table \ref{tab:combined_results}. Gen tasks, which provided 5 or 10 in-context examples, consistently achieved substantially higher scores overall with a difference of over 10 percentage points in most models. This disparity suggests that for complex, culture-specific tasks, few-shot prompting is critical to elicit strong model performance. Notable exceptions include GPT-OSS and the Gemma 4 small models, which had extremely poor results in generation-based tasks and subpar albeit not terrible results in probability-based tasks. In the generation tasks, the evaluated Gemma 4 E2B and E4B checkpoints produced predominantly empty responses, limiting the interpretation of their scores. GPT-OSS 20B also exhibited anomalous outputs: most of the responses consisted of the repeated string “**?**”, while the remaining outputs were malformed or unrelated to the task. The cause of this behavior requires further investigation and will be examined in future work, but no evidence indicates that these issues were due to the evaluation set-up.

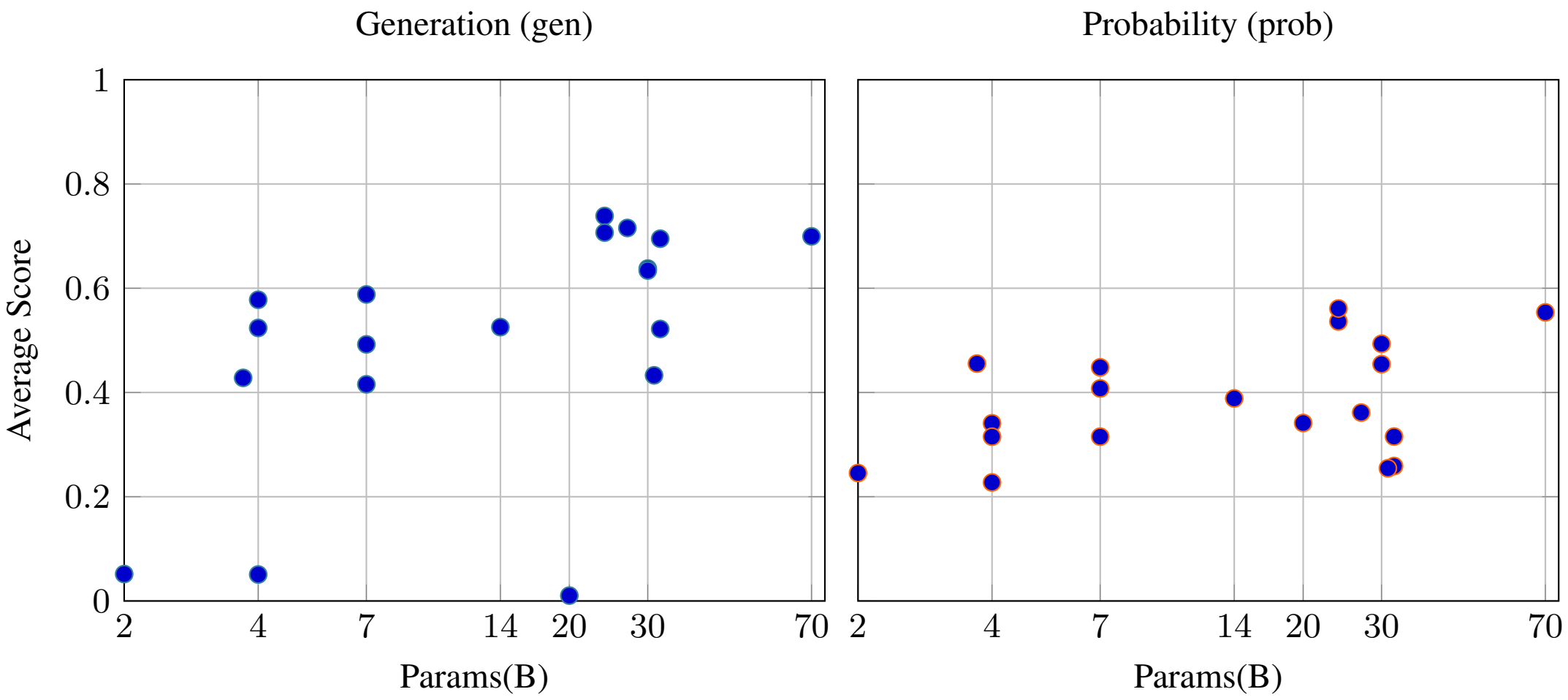
\begin{figure}[h!]
    \centering
    \begin{tikzpicture}
    \begin{groupplot}[
        group style={group size=2 by 1, horizontal sep=0.3cm},
        width=0.48\linewidth, height=6.3cm,
        xmode=log, log basis x=10, log ticks with fixed point,
        xmin=2, xmax=75, ymin=0, ymax=1,
        xtick={2,4,7,14,20,30,70},
        xlabel={Params(B)},
        ylabel={Average Score},
        grid=major,
        tick label style={font=\small},
        label style={font=\small},
        title style={font=\small},
    ]
    \nextgroupplot[title={Generation (gen)}]
    \addplot+[only marks, mark=*, mark size=2.2pt, color=cyan!55!black] coordinates {
        (3.7000,0.428004) (27.0000,0.715616) (4.0000,0.577810) (70.0000,0.699531) (32.0000,0.521812) (32.0000,0.695009) (31.0000,0.432893) (2.0000,0.051639) (4.0000,0.050768) (30.0000,0.637700) (7.0000,0.415645) (7.0000,0.492261) (14.0000,0.525413) (24.0000,0.738618) (7.0000,0.588084) (24.0000,0.706687) (4.0000,0.523781) (30.0000,0.633966) (20.0000,0.010294)
    };
    \nextgroupplot[title={Probability (prob)}, ylabel={}, yticklabels=\empty]
    \addplot+[only marks, mark=*, mark size=2.2pt, color=orange!80!red] coordinates {
        (3.7000,0.455446) (27.0000,0.361445) (4.0000,0.340909) (70.0000,0.553706) (32.0000,0.258877) (32.0000,0.315409) (31.0000,0.254610) (2.0000,0.245460) (4.0000,0.227392) (30.0000,0.454345) (7.0000,0.315259) (7.0000,0.448233) (14.0000,0.388592) (24.0000,0.536217) (7.0000,0.407770) (24.0000,0.561066) (4.0000,0.315231) (30.0000,0.493400) (20.0000,0.341444)
    };
    \end{groupplot}
    \end{tikzpicture}
    \caption{Comparative performance of language models on Generation Tasks (Gen Tasks) and Probabilistic Tasks (Prob Tasks), by the number of parameters (Params) for each model.}
    \label{fig:gen_prob_performance}
\end{figure}

Figure \ref{fig:gen_prob_performance} shows that for both categories of tasks a subtle scaling effect \cite{kaplanScalingLawsNeural2020} is observed: when we exclude the outliers listed above from the analysis, larger models generally achieve some of the best performances, mid-sized models achieve average to good performance, and smaller models achieve the worst performance. However, among mid-sized models, the Mistral-Small and Magistral-Small perform well above average for their category in both areas, demonstrating a greater command of Brazilian expressions and Portuguese grammar than most. Finally, we observed that, among the smaller models, those specifically trained in Portuguese (Tucano and Sabiá) showed consistent performance, on average, despite their targeted training and the positive results reported in existing benchmarks for the Portuguese language. This may indicate that language-specific training is less effective if the assessment is performed using mostly translated datasets. However, the stable performance across different tasks of varying types and themes demonstrates the positive effect of specialized training and the potential that such models hold.

Regarding the specific tasks, we noted that the models generally presented difficulties, especially in grammar tasks pt\_conjugate (conjugate a given verb including its necessary pronouns), pt\_der\_type\_gen (recognize the type of linguistic derivation that formed a given word) and pt\_morfema (segment a given word into its constituent morphemes), and in the linguistic variation task br\_region\_exp\_gen (identify the Brazilian region of origin of a given popular expression). These tasks are challenging due to their requirement of extremely specific cultural knowledge (as is the case with br\_region\_expr\_gen) and grammar knowledge (remaining tasks).

\begin{table*}[h!]
\centering
\resizebox{\textwidth}{!}{%
\begin{tabular}{@{}lccccccc@{}}
\toprule
&&\multicolumn{3}{c}{\textbf{Gen Tasks}} & \multicolumn{3}{c}{\textbf{Prob Tasks}}\\
\cmidrule(lr){3-5}
\cmidrule(lr){6-8}
\textbf{Model Name} & \textbf{Params(B)} & \textbf{Avg. Lang. Var. Score} & \textbf{Avg. Gram. Score} & \textbf{Overall Score} & \textbf{Avg. Lang. Var. Score} & \textbf{Avg. Gram. Score} & \textbf{Overall Score} \\
\midrule
        Llama 3.1 70B-Instruct NVFP4 & 70 & 48.80\% & 70.40\% & 59.60\% & 37.03\% & 58.76\% & 47.89\% \\
        Olmo 3.1 32B-Instruct & 32 & 28.74\% & 52.36\% & 40.55\% & 26.48\% & 26.63\% & 26.55\% \\
        DeepSeek R1 Distill Qwen 32B & 32 & 45.76\% & 69.60\% & 57.68\% & 28.96\% & 35.84\% & 32.40\% \\
        Gemma 4 31B-it & 31 & 24.83\% & 38.75\% & 31.79\% & 21.57\% & 21.74\% & 21.65\% \\
        Granite 4.0 h small & 30 & 48.03\% & 62.67\% & 55.35\% & 30.94\% & 48.61\% & 39.77\% \\
        Nemotron Cascade 2 30B-A3B & 30 & 39.77\% & 60.57\% & 50.17\% & 34.68\% & 47.79\% & 41.23\% \\
\midrule
        Qwen3.5 27B & 27 & 51.59\% & 73.50\% & 62.54\% & 34.57\% & 42.34\% & 38.45\% \\
        Magistral Small 2509 & 24 & 55.48\% & 73.53\% & 64.50\% & 37.54\% & 57.78\% & 47.66\% \\
        Mistral Small 24B Instruct 2501 & 24 & 51.31\% & 70.25\% & 60.78\% & 36.43\% & 59.41\% & 47.92\% \\
        GPT OSS 20B & 20 & 0.00\% & 1.03\% & 0.51\% & 25.16\% & 27.85\% & 26.51\% \\
        Phi-4 & 14 & 49.77\% & 51.50\% & 50.64\% & 31.52\% & 42.38\% & 36.95\% \\
\midrule
        Granite 4.0 h tiny & 7 & 35.86\% & 33.72\% & 34.79\% & 28.59\% & 25.50\% & 27.04\% \\
        Sabia 7B & 7 & 34.43\% & 48.00\% & 41.22\% & 29.78\% & 44.76\% & 37.27\% \\
        Mistral 7B Instruct v0.3 & 7 & 38.91\% & 56.64\% & 47.77\% & 30.72\% & 40.65\% & 35.69\% \\
\midrule
        Gemma 4 E4B & 4 & 0.00\% & 5.09\% & 2.54\% & 22.19\% & 17.20\% & 19.70\% \\
        Nemotron 3 Nano 4B-BF16 & 4 & 28.81\% & 50.72\% & 39.76\% & 24.50\% & 30.94\% & 27.72\% \\
        Qwen3.5 4B & 4 & 37.01\% & 57.78\% & 47.39\% & 25.38\% & 37.00\% & 31.19\% \\
        Tucano2 Qwen 3.7B-Instruct & 3.7 & 35.43\% & 37.92\% & 36.67\% & 26.53\% & 42.89\% & 34.71\% \\
        Gemma 4 E2B& 2 & 0.15\% & 5.14\% & 2.64\% & 23.37\% & 18.72\% & 21.04\% \\
\bottomrule
\end{tabular}%
}
\par\smallskip
\caption{Comparative performance of language models across Generation Tasks (Gen Tasks) and Probabilistic Tasks (Prob Tasks). For each task type, the table presents the Average Language Variation Score, Average Grammar Score, and an Overall Score, alongside the model's name and number of parameters (Params).}
\label{tab:combined_results}
\end{table*}

% \begin{figure}[!h]
%     \centering
% \includegraphics[width=0.9\linewidth]{images/gen_prob_performance.png}
%     \caption{Comparative performance of language models on Generation Tasks (Gen Tasks) and Probabilistic Tasks (Prob Tasks), by the number of parameters (Params) for each model.}
% \label{fig:gen_prob_performance}
% \end{figure}

\section{Conclusion}
The creation of a benchmark containing tasks that specifically address Brazilian concepts becomes essential as the advancement of LLMs reveals their limitations in contexts that diverge from the priorities of the international and mostly anglophone-centered companies that have historically developed them. Benchmarks like \textit{Brasilidade}, designed specifically for countries that speak lower-resource languages, help direct the attention of researchers toward the development of sovereign technologies capable of understanding linguistic, cultural, social, and educational biases and nuances that have so far been ignored or only minimally explored.

Our results reveal a slight scaling behavior, which further underlines the correlation between results and the availability of computational and thus monetary resources. Specific models for Portuguese show average results, despite having performed well above average in translated benchmarks, indicating that non-native evaluation tools can be misleading. Finally, the most challenging tasks highlight the importance of benchmarks that not only impose increasingly difficult challenges on the models being evaluated, but also standardize the entire workflow, so that the assessment of specific knowledge and skills is clear and segmented, thus allowing for the identification and mitigation of limitations.

\section{Limitations}

In this work we have not presented any tasks, datasets or results related to three of the four domains of Brasilidade, despite introducing the full benchmark's methodology. Only open-source models were evaluated, and models such as OpenAI GPT, Google Gemini, Claude Sonnet/Opus and others created for commercial purposes were not evaluated here. For few-shot tasks, we have not created a function that purposely presents examples with label-variations, but rather use a 'take first' approach. In open-ended tasks we do not use an LLM-as-a-judge in our evaluation pipeline. However, these changes are in our roadmap for future works. Despite referencing related benchmarks in section 2, we do not compare them with ours.

\section*{Acknowledgments}
This work was developed within the scope of the SoberanIA initiative, a Brazilian effort initiated in the state of Piauí, focused on ethically sourced, auditable data and technological sovereignty.

%Bibliography
\bibliographystyle{unsrt}
\bibliography{references}

\appendix

\section{Datasets Summary}
\label{sec:appendix_A}

The methodology for producing the datasets that comprise our evaluation suite was conducted by a team of computational linguists. The workflow began with the automated extraction of raw data (\textit{Scraping and/or OCR}), right after a source selection (\textit{Human curatorship}) or, in some cases such as in grammar or local culture tests, the content was entirely written by the team (\textit{Human production}). The raw data was subjected to a formatting and organization stage (\textit{Processing -- cleaning and structuring}) then transformation into standardized evaluation formats such as open-ended and multiple-choice questions. To scale the project, scripts and language model APIs were used both for augmenting examples (\textit{AI-assisted generation}) and for classifying them (\textit{AI tagging}). To control output quality, any process involving AI-assisted generation was mandatorily subjected to intermediate validations, culminating in a final audit (\textit{Human annotated and/or reviewed}), a stage where evaluators reviewed , annotated, and corrected the data based on guidelines for each dataset. Table \ref{table_datasets} shows the exact steps followed for the construction of each dataset of the first domain.

\DeclareUrlCommand\id{\urlstyle{tt}}
\begin{table}[h!]
\centering
\footnotesize
\setlength{\tabcolsep}{10pt}
\begin{tabular}{@{}>{\raggedright\arraybackslash}p{3.3cm}ccccccc@{}}
\hline
\textbf{Datasets} & \textbf{a} & \textbf{b} & \textbf{c} & \textbf{d} & \textbf{e} & \textbf{f} & \textbf{g} \\ \hline
\id{eval_coloquialismos_br_mcq_def_para_expr_balanc_pos_tagging}                      & \checkmark & \checkmark & –   & \checkmark & –   & –   & \checkmark \\
\hline
\id{eval_coloquialismos_br_mcq_def_para_expr_balanc_regiao_principal}    & \checkmark & \checkmark & –   & \checkmark & –   & –   & \checkmark \\
\hline
\id{eval_coloquialismos_br_mcq_expr_para_def_balanc_regiao_principal}    & \checkmark & \checkmark & –   & \checkmark & –   & –   & \checkmark \\
\hline
\id{conjugacoes_portugues_qa_evaluation_reduced}    & \checkmark & –   & \checkmark & \checkmark & –   & –   & –   \\
\hline
\id{formacao_morfologica}                   & –   & \checkmark & –   & –   & –   & –   & \checkmark \\
\hline
\id{grammar_bench}                          & –   & \checkmark & –   & –   & –   & –   & \checkmark \\
\hline
\id{graus_aumentativo_substantivo_qa}       & –   & \checkmark & –   & \checkmark & –   & –   & \checkmark \\
\hline
\id{graus_diminutivo_substantivo_qa}        & –   & \checkmark & –   & \checkmark & –   & –   & \checkmark \\
\hline
\end{tabular}
\caption{Mapping matrix for the methodological stages of Human curatorship (a), Human production (b), Scraping and/or OCR (c), Text processing\ (cleaning and structuring) (d), AI-assisted generation (e), AI tagging (f), Human annotated and/or reviewed (g) applied to the construction of each dataset within the evaluation ecosystem.}
\label{table_datasets}
\end{table}

\subsection{Linguistic Variation (Coloquialismos)}
All of the Linguistic Variation tasks were built using the \textit{Coloquialismos} family of datasets. The original dataset \textit{coloquialismos\_br} was constructed manually by our team of linguists using data from multiple human sources from the internet and compiled into a structured format containing fields with expressions, definitions, region of origin, morphological classification and whether the expression is pejorative/derogatory or not. All of this information was carefully reviewed before the final version was completed and different combinations of those fields were then used to create multiple datasets, which were in turn used to build these tasks. Since one of the main purposes of the Brasilidade benchmark is to expose the limitations of LLMs on specific knowledge of Brazil, an effort was made to balance the relevant fields as much as possible for each dataset so that all important variations are equally represented. For example, the tasks \textit{br\_region\_exp\_gen} and \textit{br\_region\_exp\_prob} used a version of the original \textit{coloquialismos\_br} dataset that has a balanced distribution of Brazilian regions and alternatives for the answer. Alternatives were also added so that these tasks could be created as classification by generation rather than direct generation tasks due to the intrinsic ambiguity of regional expressions and definitions. Below are examples of entries for each dataset in JSONL format.

\begin{listing}[!h]
\begin{minted}{json}
{
"expressao": "\"bico doce\"",
"definicao": "Quem beija muito ou é muito beijado.",
"exemplo_de_uso": "\"Aquele ali é bico doce.\"",
"regiao_principal": "Geral - NE",
"pos_tagging": "ADJ",
"observacoes": null,
"pejorativo": false,
"uso": "em_uso"
}
\end{minted}   
    \caption{JSON data entry of the original coloquialismos\_br dataset.}
    \label{fig:json_code1}
\end{listing}

\begin{listing}[!h]
\begin{minted}{json}
{
"id": "expr_def_1044",
"pergunta": "Dentre as opções a seguir, qual o significado correto para a expressão <\"bico doce\">?",
"alternativas": ["Quem faz birra ou manha.", "Quem beija muito ou é muito beijado.", "Pessoa que tem um comportamento instável.", "Algo considerado \"mais ou menos\", mediano."],
"resposta_correta": 1,
"resposta_correta_texto": "Quem beija muito ou é muito beijado.",
"expressao_original": "\"bico doce\"",
"pos_tagging": "ADJ",
"regiao_principal": "Geral - NE",
"exemplo_de_uso": "\"Aquele ali é bico doce.\"",
"pejorativo": false,
"observacoes": null
}
\end{minted}   
    \caption{JSON data entry for both eval\_coloquialismos\_br\_mcq\_expr\_para\_def\_balanc\_regiao\_principal and eval\_coloquialismos\_br\_mcq\_expr\_para\_def\_balanc\_pos\_tagging.}
    \label{fig:json_code2}
\end{listing}

\begin{listing}[!h]
\begin{minted}{json}
{
"id": "def_expr_1044",
"pergunta": "Dentre as opções a seguir, qual expressão é corretamente representada pela seguinte definição: <Quem beija muito ou é muito beijado.>?",
"alternativas": ["\"patolino\"", "\"bico doce\"", "\"ferrado\"", "\"capiongo\""],
"resposta_correta": 1,
"resposta_correta_texto": "\"bico doce\"",
"definicao_original": "Quem beija muito ou é muito beijado.",
"pos_tagging": "ADJ",
"regiao_principal": "Geral - NE",
"exemplo_de_uso": "\"Aquele ali é bico doce.\"",
"pejorativo": false,
"observacoes": null
}
\end{minted}   
    \caption{JSON data entry for eval\_coloquialismos\_br\_mcq\_def\_para\_expr\_balanc\_regiao\_principal.}
    \label{fig:json_code3}
\end{listing}

\subsection{Grammar}
There are five datasets that compose the tasks in the grammar subgroup. The \textit{conjugacoes\_portugues\_qa\_evaluation} dataset (listing \ref{fig:json_code4}) was constructed from the scraping of structured verb-conjugation data and a subsequent generation of Q\&A pairs on these raw data using fixed templates; \textit{formacao\_morfologica} (listing \ref{fig:json_code6}) was manually compiled by our team of linguists with different cases of word formation (whether by derivation, by composition, or by both), segmenting the terms into morphemes and explaining them; \textit{grammar\_bench} (listing \ref{fig:json_code7})was manually built by the linguistics team with common errors made by speakers of Portuguese (along with respective error categories and possible corrections); \textit{graus\_aumentativo\_substantivo\_qa} (listing \ref{fig:json_code8}) and \textit{graus\_diminutivo\_substantivo\_qa} (listing \ref{fig:json_code9}) were also both manually compiled and present possible augmentative and diminutive forms, respectively, for a selection of nouns in Portuguese. Below are example of entries for each dataset in a JSONL format.

\begin{listing}[!h]
\begin{minted}{json}
{
"question": "Qual é a segunda pessoa do singular do imperativo negativo com pronome oblíquo do verbo desbotar?",
"answer": "não o desbotes tu"
}
\end{minted}   
    \caption{Data entry part of the conjugacoes\_portugues\_qa\_evaluation dataset in the grammar subgroup.}
    \label{fig:json_code4}
\end{listing}

\begin{listing}[!h]
\begin{minted}{json}
{
"termo": "sedento",
"classe": "adjetivo",
"derivacao": "Derivação sufixal",
"composicao": null,
"observacao": null,
"morfema_1": "sed-",
"significado_1": "Relativo a sede (sede de beber); desejo intenso de líquidos.",
"classificacao_1": "radical",
"morfema_2": "-ento",
"significado_2": "marca estado, condição ou aquele que sente/possui algo.",
"classificacao_2": "sufixo",
"morfema_3": null,
"significado_3": null,
"classificacao_3": null,
"morfema_4": null,
"significado_4": null,
"classificacao_4": null,
"morfema_5": null,
"significado_5": null,
"classificacao_5": null
}
\end{minted}   
    \caption{Data entry part of the formacao\_morfologica dataset in the grammar subgroup.}
    \label{fig:json_code6}
\end{listing}

\begin{listing}[!h]
\begin{minted}{json}
{
"frase_incorreta": "Os cachorro correu atrás do gato.",
"frase_corrigida": "Os cachorros correram atrás do gato.",
"explicacao": "O verbo \"correr\" deve estar na terceira pessoa do plural (\"correram\") para concordar com o sujeito plural \"Os cachorros\".",
"categoria": "CONCORDANCIA_VERBAL"
}
\end{minted}   
    \caption{Data entry part of the grammar\_bench dataset in the grammar subgroup.}
    \label{fig:json_code7}
\end{listing}

\begin{listing}[!h]
\begin{minted}{json}
{
"id": 148,
"palavra": "poema",
"pergunta": "Dentre as opções a seguir, qual é a forma correta do grau aumentativo da palavra 'poema'?",
"grau": "aumentativo",
"alternativa_a": "poemedo",
"alternativa_a_correta": false,
"alternativa_b": "poemazão",
"alternativa_b_correta": false,
"alternativa_c": "poemão",
"alternativa_c_correta": true,
"alternativa_d": "poemaço",
"alternativa_d_correta": false,
"alternativa_e": "poemonha",
"alternativa_e_correta": false,
"resposta_correta": "poemão"
}
\end{minted}   
    \caption{Data entry part of the graus\_aumentativo\_substantivo\_qa dataset in the grammar subgroup.}
    \label{fig:json_code8}
\end{listing}

\begin{listing}[!h]
\begin{minted}{json}
{
"id": 148,
"palavra": "poema",
"pergunta": "Dentre as opções a seguir, qual é a forma correta do grau diminutivo da palavra 'poema'?",
"grau": "diminutivo",
"alternativa_a": "poemeco",
"alternativa_a_correta": false,
"alternativa_b": "poemispo",
"alternativa_b_correta": false,
"alternativa_c": "poemazinho",
"alternativa_c_correta": false,
"alternativa_d": "poemoide",
"alternativa_d_correta": false,
"alternativa_e": "poeminha",
"alternativa_e_correta": true,
"resposta_correta": "poeminha"
}
\end{minted}   
    \caption{Data entry part of the graus\_diminutivo\_substantivo\_qa dataset in the grammar subgroup.}
    \label{fig:json_code9}
\end{listing}

\section{Tasks Summary}
\label{sec:appendix_B}

Table \ref{tab:tasks} shows a list of all tasks of the first domain. It includes descriptions of task, subdomain, base YAML and dataset used.

The Base YAML column refers to the task subtype as per Section \ref{sec:task_types_subtypes}. Each subtype has their own configuration file and function for processing results inside the framework LM Evaluation Harness (e.g. the subtype Open-Ended Q\&A Generation uses the \textit{qa\_gen} base configuration file, Direct Generation uses the \textit{direct\_gen} base configuration file, etc). Each base configuration file includes the general parameters and arguments necessary to build a task of each respective subtype (e.g. metrics, process results functions, number of few-shots, maximum token count, etc). The specific configuration files of each task include arguments such as dataset path, instructions and other parameters that differentiate this task from the others. Along with all of the specific configuration files of the tasks that will be made public, we are also releasing the base configuration files, and the process\_docs, process\_results and shared\_utils scripts necessary to run them. Runtime instructions will also be found on that same folder.

\DeclareUrlCommand\id{\urlstyle{tt}}
\clearpage
\onecolumn
\begin{longtable}{@{}>{\raggedright\arraybackslash}p{2.5cm}              >{\raggedright\arraybackslash}p{5.2cm}                    >{\raggedright\arraybackslash}p{1.6cm}                    >{\raggedright\arraybackslash}p{1.8cm}                    >{\raggedright\arraybackslash}p{3.2cm}@{}}
\caption{Tasks in the Linguistic Variation and Grammar domains of the Brasilidade benchmark, with the base task configuration and the source dataset for each task.}
\label{tab:tasks} \\
\hline
\textbf{Task} & \textbf{Description} & \textbf{Domain} & \textbf{Base YAML} & \textbf{Dataset} \\
  \hline
  \endfirsthead
  \hline
  \textbf{Task} & \textbf{Description} & \textbf{Domain} & \textbf{Base YAML} & \textbf{Dataset} \\
  \hline
  \endhead
  \hline
  \endfoot
  \hline
  \endlastfoot
  \id{br_def_expr_prob} & Given the definition of a Brazilian regional expression, choose the option containing the expression being defined. & Linguistic Variation & \id{var_class_prob} & \id{eval_coloquialismos_br_mcq_def_para_expr_balanc_regiao_principal} \\
  \id{br_def_expr_gen} & Given the definition of a Brazilian regional expression, choose the option containing the expression being defined. & Linguistic Variation & \id{var_class_gen} & \id{eval_coloquialismos_br_mcq_def_para_expr_balanc_regiao_principal} \\
  \id{br_expr_def_prob} & Given a Brazilian regional expression, choose the option containing its definition. & Linguistic Variation & \id{var_class_prob} & \id{eval_coloquialismos_br_mcq_expr_para_def_balanc_regiao_principal} \\
  \id{br_expr_def_gen} & Given a Brazilian regional expression, choose the option containing its definition. & Linguistic Variation & \id{var_class_gen} & \id{eval_coloquialismos_br_mcq_expr_para_def_balanc_regiao_principal} \\
  \id{br_region_exp_prob} & Choose the Brazilian region where an expression originated from a list of options. & Linguistic Variation & \id{fix_class_prob} & \id{eval_coloquialismos_br_mcq_def_para_expr_balanc_regiao_principal} \\
  \id{br_region_exp_gen} & Choose the Brazilian region where an expression originated from a list of options. & Linguistic Variation & \id{fix_class_gen} & \id{eval_coloquialismos_br_mcq_def_para_expr_balanc_regiao_principal} \\
  \id{pt_aum_prob} & Choose which option is an augmentative form of a given word. & Grammar & \id{var_class_prob} & \id{graus_aumentativo_substantivo_qa} \\
  \id{pt_aum_gen} & Choose which option is an augmentative form of a given word. & Grammar & \id{var_class_gen} & \id{graus_aumentativo_substantivo_qa} \\
  \id{pt_category_prob} & Classify the grammatical error present in an incorrect sentence using a list of options. & Grammar & \id{fix_class_prob} & \id{grammar_bench} \\
  \id{pt_category_gen} & Classify the grammatical error present in an incorrect sentence using a list of options. & Grammar & \id{fix_class_gen} & \id{grammar_bench} \\
  \id{pt_comp_type_prob} & Choose the type of morphological composition used to form a given word. & Grammar & \id{fix_class_prob} & \id{formacao_morfologica} \\
  \id{pt_comp_type_gen} & Choose the type of morphological composition used to form a given word. & Grammar & \id{fix_class_gen} & \id{formacao_morfologica} \\
  \id{pt_conjugate} & Generate the requested conjugated form of a verb. & Grammar & \id{direct_gen} & \id{conjugacoes_portugues_qa_evaluation_reduced} \\
  \id{pt_correct} & Correct grammatically incorrect sentences in Portuguese. & Grammar & \id{direct_gen} & \id{grammar_bench} \\
  \id{pt_count_morf_prob} & State how many morphemes, from 1 to 5, make up a given word. & Grammar & \id{fix_class_prob} & \id{formacao_morfologica} \\
  \id{pt_count_morf_gen} & State how many morphemes, from 1 to 5, make up a given word. & Grammar & \id{fix_class_gen} & \id{formacao_morfologica} \\
  \id{pt_der_comp_prob} & Determine whether a word was formed through derivation or composition. & Grammar & \id{fix_class_prob} & \id{formacao_morfologica} \\
  \id{pt_der_comp_gen} & Determine whether a word was formed through derivation or composition. & Grammar & \id{fix_class_gen} & \id{formacao_morfologica} \\
  \id{pt_der_type_prob} & Choose the type of derivation used to form a given word. & Grammar & \id{fix_class_prob} & \id{formacao_morfologica} \\
  \id{pt_der_type_gen} & Choose the type of derivation used to form a given word. & Grammar & \id{fix_class_gen} & \id{formacao_morfologica} \\
  \id{pt_dim_prob} & Choose which option is a diminutive form of a given word. & Grammar & \id{var_class_prob} & \id{graus_diminutivo_substantivo_qa} \\
  \id{pt_dim_gen} & Choose which option is a diminutive form of a given word. & Grammar & \id{var_class_gen} & \id{graus_diminutivo_substantivo_qa} \\
  \id{pt_expr_class_prob} & Grammatically classify a Brazilian regional expression using a list of options, such as verb, adjective, interjection, etc. & Grammar & \id{fix_class_prob} & \id{eval_coloquialismos_br_mcq_def_para_expr_balanc_pos_tagging} \\
  \id{pt_expr_class_gen} & Grammatically classify a Brazilian regional expression using a list of options, such as verb, adjective, interjection, etc. & Grammar & \id{fix_class_gen} & \id{eval_coloquialismos_br_mcq_def_para_expr_balanc_pos_tagging} \\
  \id{pt_morfema} & Given a word, segment it into its constituent morphemes. & Grammar & \id{direct_gen} & \id{formacao_morfologica} \\
  \id{pt_word_class_prob} & Morphologically classify a word, such as verb, noun, adjective, or adverb. & Grammar & \id{fix_class_prob} & \id{formacao_morfologica} \\
  \id{pt_word_class_gen} & Morphologically classify a word, such as verb, noun, adjective, or adverb. & Grammar & \id{fix_class_gen} & \id{formacao_morfologica} \\
\end{longtable}
\twocolumn

\end{document}